\documentclass{article}
\usepackage[utf8]{inputenc}
\usepackage{graphicx}
\usepackage{caption}
\usepackage{subfig}
\usepackage{amsmath}
\usepackage{soul}
\usepackage{booktabs}   
\usepackage{booktabs, multicol, multirow}
\usepackage{authblk}

\title{Convolutional neural networks and multi-threshold analysis for contamination detection in the apparel industry}

\author[1]{Marco Boresta}
\author[2]{Tommaso Colombo}
\author[3]{Alberto De Santis}

\affil[1]{National Research Coucil, CNR-IASI, Rome, Italy}
\affil[2]{aHead Research ETS}
\affil[3]{Department of Computer, Control and Management Engineering A. Ruberti, Sapienza University of Rome, 00185 Rome, Italy}

\date{\today}

\begin{document}
\maketitle

\noindent\textbf{keywords}{
quality control; contamination detection; convolutional neural networks; image processing; X-ray images}

\section{Abstract}
Quality control of apparel items is mandatory in modern textile industry, as consumer’s awareness and expectations about the highest possible standard is constantly increasing in favor of sustainable and ethical textile products. Such a level of quality is achieved by checking the product throughout its life cycle, from raw materials to boxed stock. Checks may include color shading tests, fasteners fatigue tests, fabric weigh tests, contamination tests, etc. This work deals specifically with the automatic detection of contaminations given by small parts in the finished product such as raw material like little stones and plastic bits or materials from the construction process, like a whole needle or a clip.
Identification is performed by a two-level processing of X-ray images of the items: in the first, a multi-threshold analysis recognizes the contaminations by gray level and shape attributes; the second level consists of a deep learning classifier that has been trained to distinguish between true positives and false positives.
The automatic detector was successfully deployed in an actual production plant, since the results satisfy the technical specification of the process, namely a number of false negatives smaller than $3\%$ and a number of false positives smaller than $15\%$. 

\section{Introduction}
\label{introduction}
Apparel manufacturing is the result of a number of cooperating activities over a variety of different raw materials, product types, supply chains and production lines, production volumes, retail markets, technologies. For a full customer’s satisfaction, quality control must be practiced at any level, from the sourcing raw materials to the boxed-garment. Indeed, textile products are assessed for quality in the preproduction phase, during production, and with a final inspection after completion. Along a production line, product quality depends on many attributes like the standard of fibers, yarns, fabric construction, colour shade, surface designs, fasteners and other accessories reliability. This level of quality control includes also the checking for contaminations, which is the focus of this work. The finished product may satisfy the requirements according the previously mentioned attributes, but may well show clues of all the production activities: there may be found clips and needles stuck into the padding, needle and button tips, little stones (pebbles), also construction parts like grommets or zip pullers. Therefore, the task to be accomplished consists in detecting these kinds of contaminations and to route the checked item to a controlling room where the contamination will be removed. 

Data are 8-bit gray level images obtained by X-ray scanners. In short, the item image is segmented to separate different objects of interest from the background. Most of these objects are just artefacts produced by the scanner signal quantization (8-bit), others are part of the item (buttons, zippers, drawstrings,  toggle-sliders, …). Save for needles and clips, contaminations are usually objects of a very small size, very similar to scanner artefacts. Therefore, global segmentation methods like region-growing methods [1, 2, 3], fail to isolate the contamination. In fact, on a sample of about 590 X-ray images that constituted the available training set, it turned out that the most reliable way to isolate a contamination was to analyze the images binarized at different thresholds of increasing magnitude.  Moreover, a set of three morphological properties was found suitable to adequately characterize the contaminations shape: area, solidity and aspect ratio. Solidity is the ratio between an object area and the area of its convex hull, while the aspect ratio was defined as the ratio between the lengths of the major and minor axes of the object (referred to the ellipse that has the same normalized second central moments as the object). These parameters allow to distinguish the contamination from other detected objects. Furthermore, it was also found that the true contaminations have different behaviors on increasing threshold segmentations than most of the false contaminations, and this enables a further selection between real defects and false alarms. 

\begin{figure}[ht]
    \centering
    \subfloat[TP1]{{\includegraphics[width=5cm]{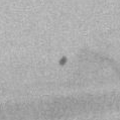} }}%
    \qquad
    \subfloat[drawstring (coulisse)]{{\includegraphics[width=5cm]{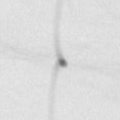} }}%
    \caption{(a) image of a pebble; (b) image of a drawstring .}%
    \label{coulisse}%
\end{figure}


The combination of these processes defines a multi-threshold filter (MT-Filter) that is able to identify most of the contaminations in apparel items, although with a non negligible number of false positives. A representative example of such a false positive is shown in figure \ref{coulisse}. The shape, dimension and grey level of the folded drawstring in fig.\ref{coulisse} (b) matches almost exactly those of a real contamination, like the one in fig.\ref{coulisse} (a). This is something that happens frequently when applying the MT-Filter to an X-ray image of an apparel item because the presence of seam lines as well as zip elements, broken stitching, toggles etc. Therefore, a further functional level of the detection procedure must be designed to lower the rate of the false positive detections. This is obtained by a deep learning classifier trained to distinguish between the real contaminations and the false positives. The classifier, once properly trained, is applied to the detections provided by the MT-Filter and gives a quick response to distinguish a true positive from a false alarm. 
The classifier has the structure of a Convolutional Neural Network (CNN-Classifier). 
{We like to stress that in the quality control of textile products the procedures are obtained as an \textit{ad hoc} adaptation of existing methods, usually more than one, to the various case studies. As it will be clear in the next Section of related works, many different technologies can be applied to collect data for visual inspection, and well established algorithms are employed mainly for texture analysis. The case study presented in this work is original within the textile quality control since the goal consists in checking the presence of small intrusions in a finite product by using X-ray data.}

The paper is organized as follows: in Section \ref{related_works} we describe existing research related to quality control of textile products. Section \ref{materials_methods} introduces the dataset at disposal, and describes the methods developed to perform contamination detection with a low false positive rate. In Section \ref{experiments_results} we present the experiments and the results obtained by applying the proposed approach to the dataset at disposal. Finally, in Section \ref{conclusions} some conclusions are given.

\subsection{Related works}
\label{related_works}
{This section  presents different case studies that occur in the quality control of textile products and the different metodologies borrowed from image analysis for visual data processing. It aims to offer a survey of possible applications of well established image analysis procedures to different problems arising in the quality check in the apparel industry.}
Quality control is a crucial aspect in the industrial production line. In \cite{Czimmermann2020} a recent review can be found, describing non destructive approaches for quality assessment. These approaches avail of different technologies (radiography, eddy current, thermography, ultrasounds,...) and the ones related to this work lie within the visual inspection framework. In this context, most of the methods are developed to check surface defects by texture analysis, that is the spatial distribution of pixel values in a recorded image. An interesting method can be found in \cite{Dastoorian2018}, where an Adaptive Generalized Likelihood Ratio is used to detect defects by checking whether the distribution of the observed data is significantly different from a baseline historical distribution in an adaptive manner. The co-occurrence matrices (GLCMs) are one of the most well-known and commonly
used texture features. They are statistical methods that measure the spatial relationship of grey-scale pixels: the occurrence frequency of pairs of pixels with specific values and in a specified spatial relationship in an image, given a displacement vector, is determined and then texture features such as energy, entropy, contrast, homogeneity and correlation are extracted. Application to textile quality control can be found in papers \cite{RAHEJA2013A,RAHEJA2013B}. To achieve satisfactory results, GLCM methods need to be carefully tuned according to the application considered; in paper \cite{Capizzi2015} for instance, the authors propose an GLCM based detector with a radial basis probabilistic neural network  to detect defects on fruits, and achieved $97.25\%$ detection rate with $ 2.75\%$ false alarm. 
A low computational cost method to describe local texture properties is the 
Local binary pattern method (LBP). The image is scanned by a sliding window, the center pixel grey level value is used as threshold and the window pixel are binarized according to this threshold \cite{OJALA1996}. The LBP operator is insensitive to changes
in illumination and image rotation, and it makes it a robust operator that has
been used in several defect detector applications on varied materials such as ceramics or wood. Anyway
it has better performances if combined with other methods: in \cite{ZHang2015} the authors combined the GLCM and LBP methods to extract image features to train a neural network that achieved $97.6\%$ detection rate on 90 samples from TILDA database; in \cite{Sindagi2015}, a modified LBP method to train a SVM classifier is proposed, achieving $93\%$ of accuracy on a sample of 148,905 images. 

Filter based approaches use signal filtering to determine the local frequency characteristics of the texture to compare with originals to detect defects. The most popular transformations are Gabor filtering \cite{Mak2008} and  Wavelet transform \cite{HAN2007}. These methods are usually computationally expensive, therefore their use in real time application must be carefully considered.

Deep learning is a fast growing field in computer sciences, because it can solve highly complex problems \cite{Zhao2019}. Deep learning  transforms data into complex 
representations that enable the features to be learned overcoming the requirement of complex features for a specific defect, typical of the so called traditional methods above reported. Deep learning is a data-driven artificial intelligence techniques able to successfully model deterministic rules, which are often incomprehensible to humans, and relationships between input and output, \cite{Tabernik2020,GOPALAKRISHNAN2017,Lin2019}.

As a general remark, we can say that  a huge number of methods is available for quality control of industrial production, but the ultimate method does not exist.
Each method comes with pros and cons, and the level of the performance mainly depends on the quantity and quality of  data available to the experimenter. Moreover, in most of the applications the detection needs to be in real time. As a matter of fact, all the approaches reported above require that there is an original flawless model from which the basic surface characteristics can be learned. In the textile production this is realistic for the fabric from the loom, but does not apply to finite products. In the application considered in this paper an apparel item is just passed through the X-ray scanner 
{and no particular care is taken in positioning it in the scanner, so} it can be  folded in an uncontrolled manner. In this situation, {there is no a master picture of the item to be compared to the X-ray data to check for differences of any kind.} 
Moreover the defects are just very small particles over large size images, with level and shape very similar to the background artefacts similar to noise in X-ray data. Therefore even machine learning methods do not guarantee good performances. {This case study indeed is different from those usually found in the relevant literature, and calls for a method with mixed properties}: a filter subsystem isolates the candidates for contaminations and a machine learning classifier distinguishes among these candidates the true defects from the false alarms.

\section{Materials and methods}
\label{materials_methods}
\begin{figure}[!ht]
    \includegraphics[width=380pt, height = 99.665pt] {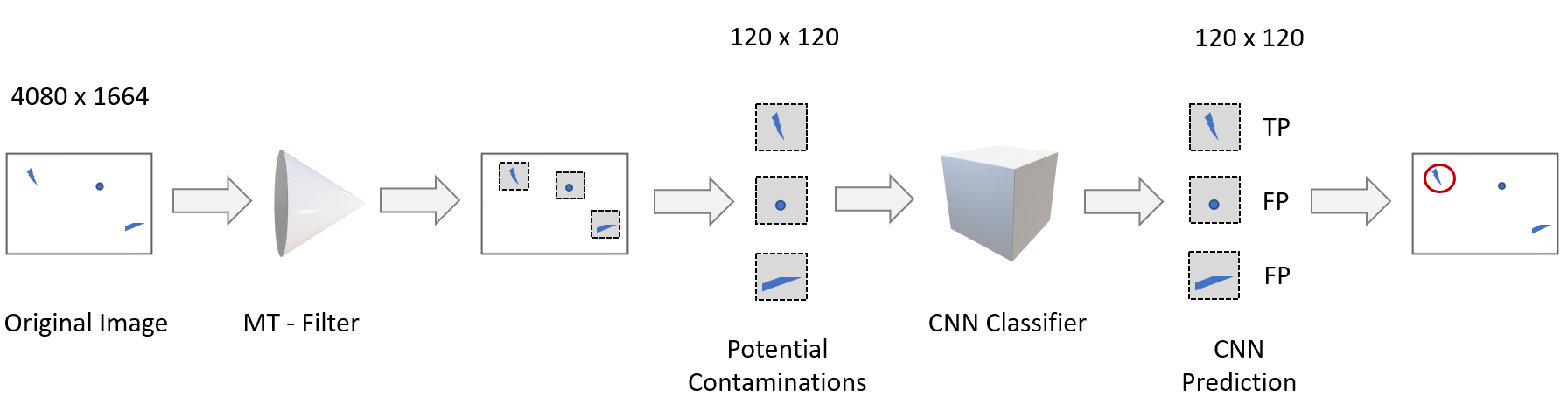}
    \caption{Scheme of the contaminations detection process}
    \label{fig:schema_metodologia}
\end{figure}

In this section the dataset and the identification procedure are described. The identification procedure is composed of two processes that cooperate to obtain a robust and efficient detection of contaminations in an apparel item, see Figure \ref{fig:schema_metodologia}. Robustness refers to a high rate of true positives and a low rate of false alarms, while efficiency is needed for a real time detection. The contaminations (intrusions) are indeed small size objects to be classified either as defects of the apparel item as compared to other objects that are parts of the item (zip, puller, buttons, drawstring) or to artefacts generated by the X-ray scanner. It is quite inefficient to feed the classifier with the whole X-ray image of the item since, due to its large size, it would result in a complex system (high number of parameters) very difficult to train to distinguish large size images that differ  because of  very small size subsets of pixels. Complexity could be reduced by a sliding window system that inputs the classifier with reduced size sub-images obtained by sliding a window over the whole image domain, but the resulting process would be time consuming and would yield to a considerably high number of False Positives. Therefore an upstream process is needed to select a finite number of instances to pass to the classifier. This is accomplished by a multi-threshold filter that efficiently segments the original image into component objects, filters out the objects by gray level, shape and density selection, and determines a small number of pixel subsets that are candidates to be a true contamination.

\subsection{The dataset}
The dataset used in this work is property of a private international company working in the apparel industry.
We had two different groups of images, one used for the MT-Filter and the other for the CNN-Classifier.

\begin{itemize}
    \item For the filter there were 11598 images with dimensions 4080 x 1664 pixels. Because of privacy reasons, we are not allowed to show images as a whole, but only some of their details. In these sub-images, components such as zip-pullers, buttons and toggles-sliders appear as dark objects. Seams, drawstrings, zip-teeth have and intermediate gray tone, very similar to artefacts produced by the item creases. Figure \ref{details_contamination}(a) shows an image detail where a needle bit contamination is present. It appears as a quite dark object of small size. In Figure  \ref{details_contamination} (b) a pebble contamination is displayed, it has a quite light gray level and very small area (it can be as small as 5 pixels).

    \begin{figure}[h]
        \centering
        \subfloat[label 1]{{\includegraphics[width=4.5cm]{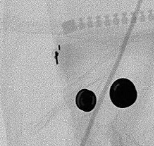} }}%
        \qquad
        \subfloat[label 2]{{\includegraphics[width=5cm]{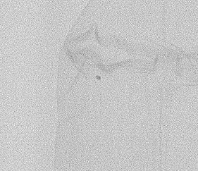} }}%
        \caption{This figure shows two typical contaminations: (a) needle bit; (b) pebble.}%
        \label{details_contamination}
    \end{figure}
    
    Of these images, 590 were used to tune the parameters of the filter, while the rest were part of the test set, as described in section \ref{filter_subsystem}.

    \item For the classifier there were 10123 images with dimensions 120 x 120 pixels, obtained as crops of the original image around either a contamination, from here on referred to as True Contamination (or TC), or around an actual part of the item, from here on referred to as False Contamination (or FC). In particular,
4988 were images of TC and 5135 were images of FC. An example of a TC and a FC image is represented in Figure \ref{coulisse}.
\end{itemize}

\subsection{The MT-Filter}
{The task consists in detecting all kinds of  contamination within given ranges of gray level and of some parameters defining the shape}. 
{To this aim, let us define a set of gray level threshold values as follows}
$$
th_k = k\,\frac{255}{24} = k\delta,\quad k=0,\ldots,23.
$$
{
Then, for any  $0\le k\le 23$ a binary image $I_{k}$ is obtained according the following rule}
\begin{equation}
I_{k}(i,j) = \begin{cases} 1, &  I(i,j) < th_{k},\label{binarule}\\
0, & \text{otherwise}
\end{cases}
\end{equation}
{ Pixels whose gray level is within the range $[0,\,th_{k})$ are labeled as white, all the others are labeled as black. 
Let us choose for instance  $k=7$ and apply rule (\ref{binarule}) to both Figure \ref{details_contamination}(a) and \ref{details_contamination}(b).}

\begin{figure}[h]
    \centering
    \subfloat[]{{\includegraphics[width=5cm]{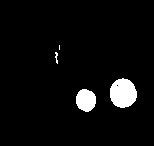} }}%
    \qquad
    \subfloat[]{{\includegraphics[width=4.8cm, height=4.75cm]{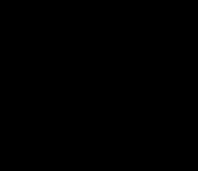} }}%
    \caption{Image binarized by \textit{$th_{7}$}: (a) binarization of Figure \ref{details_contamination}(a); (b) binarization of Figure \ref{details_contamination}(b).}%
    \label{binarized}%
\end{figure}
\begin{figure}[htbp]
    \centering
    \subfloat[]{{\includegraphics[width=5cm]{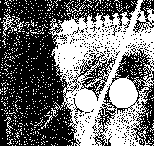} }}%
    \qquad
    \subfloat[]{{\includegraphics[width=4.8cm, height=4.75cm]{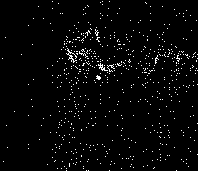} }}%
    \caption{Image binarized by \textit{$th_{16}$}: (a) binarization of Fig.\ref{details_contamination}(a); (b) binarization of Fig.\ref{details_contamination}(b).}%
    \label{fig:examplee}%
\end{figure}

\noindent 
{
The needle bit of Fig.\ref{details_contamination}(a) is clearly detected on Fig.\ref{binarized}(a) along with the two large buttons, and can be addressed by “filtering” the white objects by size and shape. On the contrary the pebble of Fig.\ref{details_contamination}(b) is not detected on Fig.\ref{binarized}(b), since it has a gray level higher than  $th_7$.}
\noindent{Now let us choose a threshold $k=16$,  suitable to detect the light gray pebble and apply it to both image details \ref{details_contamination}(a),(b) and see what happens. This time a different problem arises. Figure \ref{fig:examplee}(a) is filled with false detections while on Figure \ref{fig:examplee}(b) the pebble is clearly demarcated but it is within a cloud of artefacts, even though more sparse as compared to those on Fig. \ref{fig:examplee}(a). These artefacts can be got rid off by matched dilation/erosion morphological operations and by density filtering: the average distance of a contamination by neighbouring objects is larger as compared to that of artefacts within a cloud.}

It is clear that a single threshold can not suit all kind of contamination. Therefore, a proper set of  thresholds $[t_{k_l},\ldots,t_{k_u}]$ must be  chosen  to detect the contaminations at various gray levels, that are typical of the given textile production .

\medskip

\noindent The following steps outline the detection procedure of the contaminations.

\begin{enumerate}
    \item set $k = k_l$
    \item Binarize  by threshold $th_{k}$, according to (\ref{binarule})
    \item  For any white object compute the following quantities
        \begin{itemize}
            \item[i.]	Area, defined as the number of pixels in the object
            \item[ii.]	Ratio between the major axis and minor axis lengths (axes of the object ellissoid)
            \item[iii.]	Solidity, the ratio between the area and the area of the object convex hull
            \item[iv.]	Centroid coordinates.
        \end{itemize}
    \item	Filter out objects whose shape parameters Area, Ratio and Solidity are not within given confidence intervals and find the set \textit{O} of possible  contaminations by also\begin{itemize}
    \item[$-$] remove artefacts by matched dilation/erosion morphological operations
    \item[$-$] filter out objects whose average distance by neighbouring items is below a given threshold value $d_0$.
    \end{itemize}
    \item 	For any element $p\in O$, select a neighborhood of $I$ around the element centroid and binarize this subimage by the thresholds $th_{k}+\delta$. The size of every element increases since other pixels may join its pixels set.
    
    \item 	Distinguish a true contamination from a false detection: for any element in $O$ check the increase in area and major axis length of the corresponding object within the subimage binarized at the extended range $th_{k}+\delta$. If the increase rates of area and major axis length are below given upper bounds, then the element is labeled as a contamination. 
    
    \item set $k = k+1;$ if $k < k_u$ go to step 2
    
\end{enumerate}

\noindent Steps $1)$ to $7)$ define a selective procedure to segment  objects at different gray level values; it detects contaminations by shape filtering (step 4)) and a shape stability check over increasing threshold (step 6)). There is a number of parameters that need to be calibrated to obtain good performances
\begin{itemize}
    \item[-] the set of thresholds $[t_{k_l},\ldots,t_{k_u}]$
    \item[-] the confidence intervals for the shape parameters; they are obtained as $ \mu_{p} + 2\sigma_{p}$  with $ \mu_{p}$ the sample average and $\sigma_{p}$ the sample standard deviation
    \item[-] the size of the structuring elements for the morphological operations
    \item[-] the threshold values for the density filtering
    \item[-] the increase rate for area and major axis length in checking for false detection
\end{itemize} 

\begin{figure}[!ht]
\centering
\subfloat[]{\includegraphics[width=4.5cm]{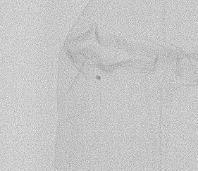}}\quad
\subfloat[]{\includegraphics[width=4.5cm]{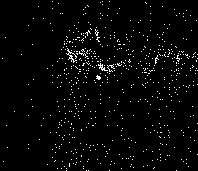}}
\\[\smallskipamount]
\subfloat[]{\includegraphics[width=4.5cm]{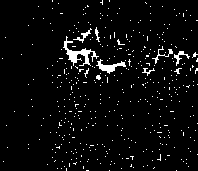}} \quad
\subfloat[]{\includegraphics[width=4.5cm]{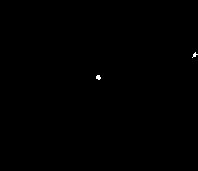}}
\\[\smallskipamount]
\subfloat[]{\includegraphics[width=4.5cm]{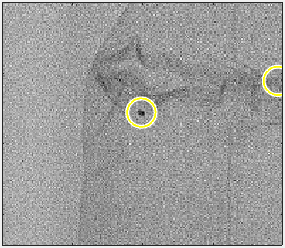}} \quad
\subfloat[]{\includegraphics[width=4.5cm]{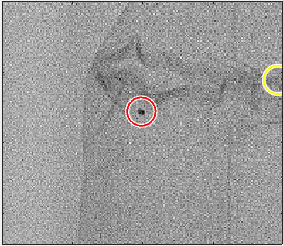}}
\caption{The various steps of the MT-Filter: (a) the original image of Fig.\ref{details_contamination}(b); (b) the binarized image with $th_{16}$; (c) the morphological processing erases some artefacts and aggregates others; (d) the shape filtering selects only the objects of panel (c) that have the shape parameters values within given intervals (this time the density check was ineffective); (e) the two objects encircled on the original picture; (f) the pebble is the detected through the check at the increased threshold value}
\label{fig:exampleee}%
\end{figure}

All these quantities can be obtained by a supervised calibration on the available training set of X-ray images of a given textile production. 
Let us show how the detection procedure works on figure \ref{details_contamination}(b). Panel $(b)$ of Figure (\ref{fig:exampleee}) shows the binarized image according to procedure (\ref{binarule}) with threshold $th_{16}$, due to the light gray level of the pebble to be identified. The pebble object is surroundend by a cloud of artefacts due to the background and other structures in X-ray image of the apparel item. Many of these can be eliminated by dilation/erosion: in panel $(c)$, the artefacts that were very close one to another have been aggregated into a larger area object that can easily eliminated by shape filtering, as can be seen on panel $(d)$, where only two objects survived. In this particular example the two objects detected were quite distant so that the density check did not modify the result. Finally checking the stability of the values (variation within given ranges)  of the shape parameters on the same object detected at the increased threshold $th_{16}+\delta$ allowed to identify the true contamination on panel $(f)$, and eliminate the false alarm due to the artefact. {Nevertheless, the MT-Filter may well generate false positives as shown in Figure (7): the true positive is detected (the leftmost red circle) along with two false alarms due to the strawstring knots, that passed all the filter checks. Therefore a classifier is required to distinguish true contaminations from artefacts.}

The shape description of the possible contamination could be in principle enriched to lower the false positive rate of the procedure, but it would imply an increase in the computational burden. It must be taken into account that the detection system must operate in nearly real time, and therefore no more than few second (typically $5$ sec) are allowed, on quite large size images. 

\begin{figure}[!ht]
\centering
\includegraphics[width=4.5cm]{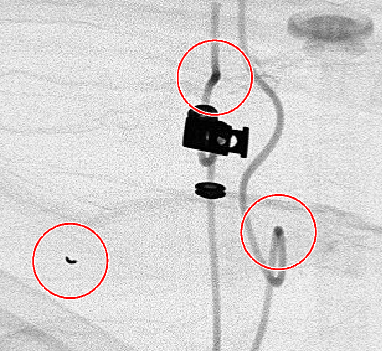}
\label{fig:false_positive_example}
\caption{False positives generated by the MT-Filter.}
\end{figure}

\subsection{Deep Convolutional Neural Networks - fundamentals}
\begin{figure}[htbp]
\centering
\includegraphics[width=12cm]{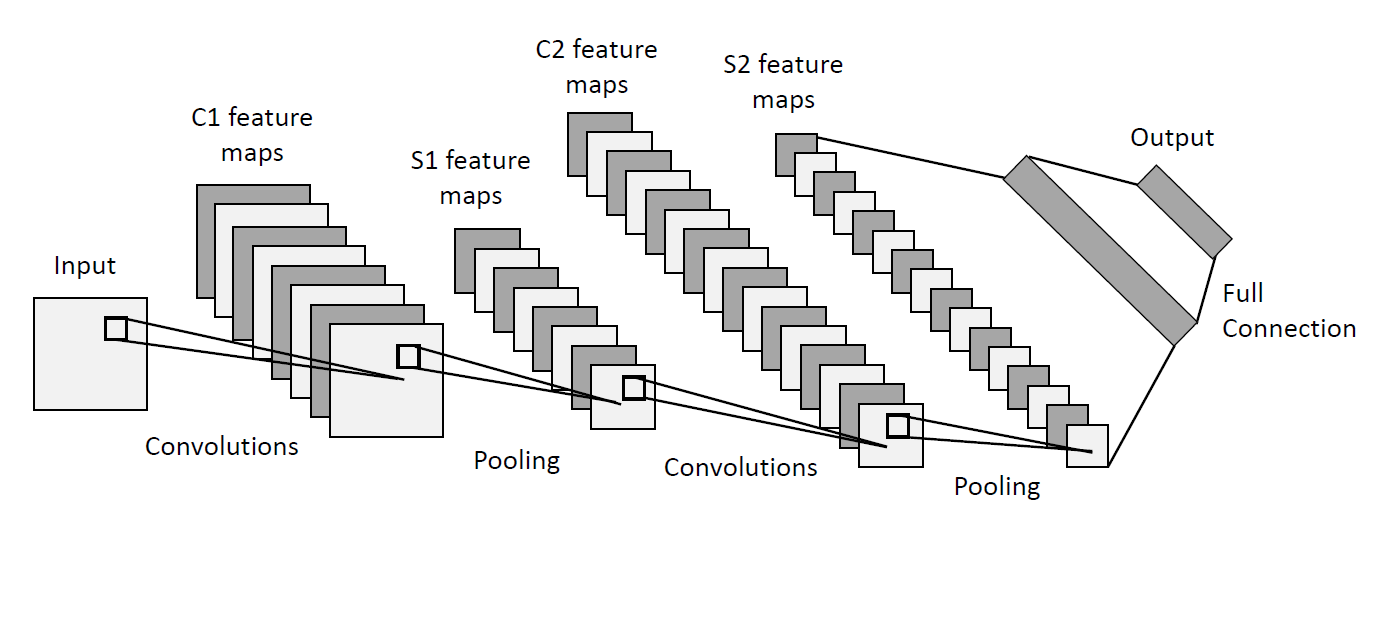}
\caption{Basic deep convolutional network architecture}
\label{CNN_architecture}
\end{figure}

As mentioned in section \ref{introduction}, the second part of the contamination detector consists of a deep learning classifier trained to distinguish between real contaminations and the false positives. The classifier is applied to the detections that are the outcome of the
MT-Filter, and is based on a Convolutional Neural Network.
\noindent
Like other types of neural networks, deep convolutional neural networks (CNN) have a layer-wise design, as shown in the example of Fig.\ref{CNN_architecture}. CNN are composed of several modules, each consisting of three different layers:

\begin{itemize}
    \item convolutional layer
    \item nonlinear layer,
    \item pooling layer

\end{itemize}

The sequence of these layers is repeated several times leading to a multi-layer structure
with very large number of layers. The last part of CNN is a conventional multi-layer
feed-forward neural network with additional dropout between its layers.

In the next subsection we briefly describe each of the components of a CNN. For more details we recommend \cite{goodfellow2016deep, lecun2010convolutional}. 
\subsubsection{Convolutional layer.}

The goal of the convolutional layer is that of extracting different sort of features from the input image. The initial layers select low level features, such as edges, shapes or colours, while the following layers are responsible for the detection of higher level features.
The output of each convolutional layer consists of several feature maps, their number corresponding to the number of filters in that layer. Each filter, responsible for the detection of a different feature, is a simple array of weights, whose value is adjusted and learnt during the training. 
The number and size of filters in each layer is an hyperparameter that is chosen in advance by the user.  

\subsubsection{Nonlinear layer.}
Each convolutional layer is followed by a nonlinear layer. This improves the approximation abilities of the network. Among several nonlinear transformations that can be used, the most common is the rectified linear unit (ReLU), that ensures faster training when used in deep convolutional networks \cite{krizhevsky2017imagenet}, basically because it mitigates the well known vanishing gradient effect during the back-propagation training in neural networks \cite{glorot2011deep}.

\subsubsection{Pooling layer}
The combination of convolutional and nonlinear layer is often followed by a pooling layer. Its job is that of reducing the size of the feature maps, in order to decrease the computational power required to process the data and to make the representation rotational and positional invariant with respect to the input \cite{lecun1990handwritten}.
Average pooling and Max pooling are two different types of pooling layer, the latter being adopted more frequently than the former. In both cases there are no weights that need to be learnt during the training phase.

\subsubsection{Dropout layer}
To overcome overfitting thus improving the generalization abilities of the network, a regularization technique known as Dropout is employed in this layer \cite{hinton2012improving}. This technique consists in randomly removing some connections during the training, with a probability determined by the dropout rate, another hyperparameter chosen in advance by the user. Dropout layers are normally removed after the training.

\subsubsection{Fully connected layer}
After several stages of convolutional and pooling layer, the feature extracted are fed to one or more fully connected layers. The last of them is then followed by a sigmoid neuron in the case of binary classification or by a softmax layer in multiclass classification.

\subsubsection{Training Procedure}

In this subsection we outline the set up procedure for the Convolutional Neural Network employed to reduce false positives. It can be described as the successive application of the following operations:
\begin{enumerate}
    \item the preparation of the data set, needed to reduce biases in the distribution of input data, i.e. the creation of a training, validation and test data sets, and the data augmentation procedure described in subsection \ref{data_set_cross_val};
    \item the choice of the metrics to evaluate the performance on the validation and test sets;
    \item the definition of the architecture of the network, i.e. the number of layers of each type (convolution, max-pooling, fully connected) and the neurons number/ kernel sizes of each layer;
    \item the identification of the parameters for the training procedure, i.e. the loss function to minimize, the optimization algorithm and its parameters (learning rate, momentum).
\end{enumerate}

We report a brief summary, in form of a pseudo-algorithm, of the whole procedure. The pseudo-algorithm will include the cross-validation procedure to choose the values of four hypothetical hyper-parameters $a,b,c,d$: the next subsections will introduce and explain all the parameters that needed to be chosen for the actual experiments.

\begin{enumerate}
    \item Given the set $\cal N$ of the available images, randomly select a test set $T_e$ of size 20\% of the original data set $\cal N$, defining with $T_r$ the set of the remaining 80\% of images.
    \item Divide the set $T_r$ of images into K folds: $T_r^1, T_r^2, \hdots, T_r^K$.
    \item Select a range of possible values for each of the hyper-parameters $a,b,c,d$;
    \item For \textit{l} iterations:
    \begin{enumerate}
        \item Randomly select one value for each hyper-parameter;
        \label{hyper_param_label}
        \item For k = 1 $\hdots$ K:
            \begin{enumerate}
                \item select $T_r^k$ as validation set, and consider the union of the other $T_r^p, p\neq k$ folds as training set.
                \item Apply data augmentation techniques on the training set, and train the CNN using the hyper-parameters selected at step  \ref{hyper_param_label}.
                \item test the CNN on the validation set $T_r^k$ and store the evaluation metrics in a list.
            \end{enumerate}
        \item Compute the average of the evaluation metrics of the K different validation sets, and store them in a table.
    \end{enumerate}
    \item Pick the combination of hyper-parameters that yields the highest average evaluation metric.
    \item Train the CNN using such combination of hyper-parameters on the whole set $\cal N$ of images and evaluate the performances on the test set $T_e$.
\end{enumerate}

In the following subsections we explain the details of some of the steps of this pseudo-algorithm.

\subsubsection{Data set preparation}
\label{data_set_cross_val}
To choose the right configuration of the hyper-parameters we implemented a random search with a K-fold cross-validation procedure, with $K=5$, that works as follows:
the training set $T_r$ is divided into $K$ folds. We run $k=1, \hdots, K$ iterations, at each of which we use $K-1$ folds as the training set, and the $k-th$ one as the validation set. 

During the training phase data augmentation techniques are used to increase the size of the dataset, therefore increasing the robustness and improving the quality of the classifier.
These techniques can be summarized as the attempt to create new data entries from existing ones by rotating, shifting, zooming in, zooming out original images, in order to generate additional artificial images and consequently prevent overfitting. At each iteration, we created some new images for each image in the training set, leaving the validation set with the original number of images.

\noindent
We then train the network with a given configuration of the hyper-parameters and we compute several metrics such as accuracy, precision, recall and $F_{\beta}$ score (see below) on the validation set.
After the last iteration we compute the average of the metrics on the $K$ different validation sets, and we store them in a table.

\subsubsection{Evaluation metrics}
\label{evalution_metrics}

As the main performance metric we chose the well known $F_{\beta}$ score. This score is an extension of the F-score or F-measure, a metric that takes into account both the precision and recall of a test's accuracy. $F_{\beta}$ is often used when the balance between precision and recall must be controlled and one wants to weigh more one than the other \cite{baeza1999modern}. To describe the way it is computed, we first need to introduce the concepts of True Positive (TP), False Positive (FP) and False Negative (FN). We define as TP the contamination in the image that is successfully detected by the classifier. If the algorithm identifies a contamination that is not in the image, we define it as a FP. Finally, FN describes a contamination that is in the image but is not detected by the algorithm. 

We can now show the formula to compute the $F_{\beta}$ score:
$$
F_{\beta} = (1+\beta^2) \, \frac{\text{precision} \cdot \text{recall}}{(\beta^2 \cdot \text{precision}) + \text{recall}}
$$
where formulae for \textit{precision} and \textit{recall} can be found in equations (\ref{precision_formula}) and (\ref{recall_formula}), respectively.
\begin{equation}
\label{precision_formula}
    precision = \frac{TP}{TP+FP}
\end{equation}

\begin{equation}
\label{recall_formula}
    recall = \frac{TP}{TP+FN}
\end{equation}
The highest possible value of $F_{\beta}$ is 1, while the worst is 0.
In the experiments, we set $\beta = 2$, since it was important for the application described in the previous sections to reduce false positives, but without sensibly impacting the performance on the false negatives. The choice of such value was ultimately driven by the request of weighting recall higher than precision.

{We conclude this section by outlining some limitations of the proposed procedure. At filter level, not all the kinds of objects that can contaminate an apparel item are detected. For example, objects of small size with a very light grey level do resemble the X-ray scanner artefacts on the image background , and therefore trying to detect these objects would give rise to a very large number of instances to pass to the classifier and a consequent unacceptable increase of the false positive rate and computation time. This limits the upper value $t_{k_u}$ that can be chosen for the multi-threshold segmentation. 
Another limitation is that some contaminations are difficult to describe with a limited number of shape parameters, like a piece of plastic string which can appear folded and knotted in an unpredictable way. At classifier level there may be a lack of a sufficient number of training examples to learn particular type of contaminations that occur rarely, like a needle that is occluded, for example, by a button, zip puller, or another item accessory.}

\section{Experiments and results}
\label{experiments_results}

In this section the MT-Filter and the CNN-Classifier procedures are applied to the considered case study. 
Subsection \ref{filter_subsystem} describes the procedure used to tune the parameters of the MT-Filter and shows its initial performance; subsection \ref{cnn_classifier} describes the hyper-parameters involved in the training of the CNN-Classifier, and presents the metrics on the test set; finally, subsection \ref{filter_cnn_combined} shows the results obtained by the combination of the filter system and the classifier, highlighting the differences from the results obtained by the only application of the filter.
\subsection{MT-Filter}
\label{filter_subsystem}

The filter subsystem is calibrated on a set of 590 X-ray images (a ground-truth data set representative of the reference textile production) that have been checked by expert personnel: on these images the various contaminations, when present, are labeled (needle, clip, plastic bit, pebble, etc.) and their coordinates are also provided. Therefore, a supervised detection procedure was first adopted on the ground-truth data set to isolate the contaminations of various kind and define their range of gray level values along with the confidence intervals for their shape parameters. Out of this first analysis, the range $[t_{k_l},\ldots,t_{k_u}]$ for the multi- threshold procedure outlined in section 4.2 and  the key information needed for the shape filtering of the detected objects at step 4) of the same procedure, are obtained. Then, steps from 1) through 7) are repeatedly applied  by a trial-error procedure for the tuning of  the parameters that characterize the data transformations building-up each step of the detection procedure: the shape and size of the structuring elements for the morphological operations, the threshold values for the density filtering, the increase factor for area and major axis length to avoid false detections. 

The tuning procedure showed that it is possible to detect all kinds of intrusions, bringing the number of false negatives to zero. 
Nevertheless, some pebbles may occur at a very light gray level (higher threshold values) and with very small size (about 5 pixels of area). They become therefore indistinguishable by a great deal of the background artefacts on the X-ray data. Having zero false negatives would result in an unacceptable rate of false positives, so that a suitable trade-off must be found between the two contrasting goals. 

The procedure tuned in the way described above was tested on a set of 11008 images, 113 of which had some contaminations and 10896 that were contaminations free. The resulting confusion matrix is the following:
$$\begin{bmatrix}
  4507&\hspace{-0.3cm}(TN)&6388&\hspace{-0.3cm}(FP) \\
   2&\hspace{-0.3cm}(FN)&111&\hspace{-0.3cm}(TP) \\
\end{bmatrix}$$
We can see that to obtain such a low false negative rate (1.77\%), the number of false positives produced is elevated, the rate being around 58\%. 

 As the filter could not be improved to lower the false positive rate, this task is demanded to the classifier. 

\subsection{CNN-Classifier}
\label{cnn_classifier}

In this subsection we describe the hyper-parameters involved in the training of the CNN-Classifier and the way they have been tuned. We analyze the role played by the \textit{class weights} and show how its values affect the various metrics on the validation sets. Finally, we present its performance on the test set.

\subsubsection{Network architecture and training parameters}

Defining the correct architecture of a deep (convolutional) neural network is not a simple task. Indeed, a trade-off between training complexity and predictive capability must often be found.
Indeed, increasing the number of layers and neurons per layer brings, theoretically, more predictive power, but also makes the training much longer and harder, therefore potentially hindering the final aim that is to find the best configuration in terms of predictive power.

The optimization algorithm that we chose is ADAM \cite{kingma2014adam}, well-known for deep learning training. ADAM's main parameters are the learning rate ($\alpha$, aka as the step size in the operations research community) and momentum $\mu$.
In general, a smaller learning rate will increase stability of the optimization procedure, but will make the training slower; the momentum parameter aims at speeding up the optimization when a consistently descent direction in the minimization of the loss function $l_f$ is found. 

All the above mentioned parameters, both network architecture and optimization algorithm ones, were chosen through a random search, where different combinations of hyper-parameters are tried, for each of which we run the k-fold-cross validation described in subsection \ref{data_set_cross_val}. 

In particular, we identified some suitable ranges of values for the following hyperparameters:

\begin{itemize}
    \item $n_c$, $n_m$, $n_f$, the number of convolution, max-pooling and fully connected layers, respectively;
    \item $\{N_k^i\}_{i=1}^{n_c}$, $\{N_m^j\}_{j=1}^{n_m}$, $\{N_f^h\}_{h=1}^{n_f}$, the size of the kernel, of the max-pooling matrix and the number of neurons of each convolution, max-pooling and fully connected layer, respectively;
    \item $\alpha$, $\mu$, the learning rate and the momentum, since the optimization algorithm of choice was the well-known ADAM algorithm. 
    \item batch size 
    \item number of epochs
    \item class weights
\end{itemize}

The first hyper-parameters are ordinary and are the subject of choice of most machine learning applications.
The last of them, the \textit{class weights}, is less common, and is normally used for dealing with unbalanced classes, along with other sampling techniques \cite{imbalance_3, imbalance_5,cacciarelli2021drives, imblearn}. 
In our application the data were almost perfectly balanced, but we had the request of weighting the recall higher than the precision, i.e. to "pay more attention" to samples belonging to the TC class. This is the why we introduced this hyper-parameters to weight instances of the two classes differently when computing the loss function, in our case the \textit{binary crossentropy}.
The parameter \textit{class weights} has the form of a tuple, where the first element represents the weight of each sample of the class FC (False Contamination) in the loss function, while the second element represents the weight of each sample of the class TC (True Contamination) in the loss function. For example, \textit{class weights} = $(1,5)$ indicates that every instance of the TC class weights 5 times more than each instance of the FC class in the computation of the binary cross-entropy.

The random search produces a table that has one row for each combination of hyper-parameters found, and one column for each metric. To choose the \textit{best} combination of hyper-parameters we look for the one that maximizes the main evaluation metric, i.e the $F_{\beta}$ score as described in \ref{evalution_metrics}

\subsubsection{CNN results}

In the following table we report the average values on the $k$ different validation sets of the $F_2$ score, $F$ score, accuracy, precision, recall and the confusion matrix for the \textit{best} combination of hyper-parameters, as described in the previous section. The value of each hyper-parameter is not reported, but we show how the various metrics are sensible to the change of the parameter \textit{class weights}. In particular, each row of the table is referred to the \textit{best} combination of hyper-parameters, with the only exception of the parameter \textit{class weights} that changes assuming four different values.

The highest entry for every metric is highlighted in bold. 

\begin{table}[htbp]
  \centering
  \caption{
  Average Evaluation Metrics on k different validation sets}
    \begin{tabular}{ccccccc}
    Class   &  \multicolumn{1}{c}{$F_2$ score} &$F$ score & \multicolumn{1}{c}{Accuracy} & \multicolumn{1}{c}{Precision} & \multicolumn{1}{c}{Recall} & Confusion \\
    weights    &  &  &  & & & matrix \\
    \midrule
    \textbf{(1, 1)}   & 0.944 &\textbf{ 0.929} & \textbf{0.928} &\textbf{0.904}&	0.955&
 $\begin{bmatrix}
    740.6       & 81 \\
    36      & 762 \\
\end{bmatrix}$
 \\
    \midrule
    \textbf{(1, 2)}   & \textbf{0.952} &0.913&	0.908&	0.854&	0.981
 & $\begin{bmatrix}
    687.4       & 134.2 \\
    15.4      & 782.6 \\
\end{bmatrix}$  \\
    \midrule
    \textbf{(1,5)}  & 0.949&	0.893&	0.883&	0.812&	0.991
& $\begin{bmatrix}
    638.4       & 183.2 \\
    7      & 791 \\
\end{bmatrix}$ \\
    \midrule
    \textbf{(1,10)}  &0.940&	0.871&	0.855&	0.777&	\textbf{0.992}
 & $\begin{bmatrix}
    593.8       & 227.8 \\
   6.4      & 791.6 \\
\end{bmatrix}$ \\

    \bottomrule
    \end{tabular}%
  \label{tab:no_noise_2}%
\end{table}%

We can see that, as the weight associated to the TC label increases, the number of False Negative decreases, along with an increase of the False Positives. This consequently affects the values of the Precision, that decreases as we move across the rows, and the Recall, that has the opposite behaviour.

Combination of weights (1,10) has the highest recall and seems to be the one that better fulfills the request of "paying attention" to the TC, i.e. to have a low number of false negatives. At the same time, it is worth noticing that a 0.01 increment in the value of the recall for the combination of weights (1,10) compared to the one of weights (1,2) comes with a price: a decrease of the precision of almost 0.08.

This suggests that there is a trade-off between these two metrics and that, depending on the application, different choices can be made. 
For the application mentioned in this paper, the combination of weights that maximizes the $F_2$ score is (1,2), that guarantees a sufficiently small number of False Positives while keeping the number of False Negatives reasonably low.

In table \ref{tab:metric_test} we report all the metrics obtained on the \textit{test set} after training the network on the whole \textit{training set} with the aforementioned best combination of hyper-parameters found.

\begin{table}[htbp]
  \centering
  \caption{
 Evaluation Metrics on the test set}
    \begin{tabular}{ccccccc}
    \multicolumn{1}{c}{$F_2$ score} &$F$ score & \multicolumn{1}{c}{Accuracy} & \multicolumn{1}{c}{Precision} & \multicolumn{1}{c}{Recall} & Confusion \\
     &  &  & & & matrix \\
    \midrule
     0.948 & 0.911 & 0.907 & 0.856&	0.974&
 $\begin{bmatrix}
    864       & 163 \\
    26      & 972 \\
\end{bmatrix}$ \\
    \bottomrule
    \end{tabular}%
  \label{tab:metric_test}%
\end{table}%

\noindent
We can see from this table that the metrics on the test set are very similar to the average of the ones on the different validation sets, thus suggesting that the training procedure produced a model with a good generalization capability.  

The experiments were performed on a computer with an Intel
Core i7-8650U processor with 1.90 GHz CPU and 16 GB of RAM,
which ran on the Windows 64-bit operating system. 
The network has been implemented in Python 3.7, using the Tensorflow/Keras framework.

\subsection{MT-Filter and CNN-Classifier combined}
\label{filter_cnn_combined}

In this section we report the results of the combination of the filter system \textit{and} the  classifier on the test set of 11008 images of size 4080 x 1664 pixels.

The two parts of the proposed approach interact in the following way:
each image is passed as input to the MT-Filter, that returns either an empty list, meaning that no potential contaminations have been detected, or a list with the coordinates of potential contaminations.

In the latter case, the image is cropped around every point identified by the coordinates, producing  sub-images of size 120 x 120 pixels.

These sub-images are then processed by the CNN-Classifier whose response labels the potential contamination either as a false positive, therefore discarded, or as a true positive, therefore kept as an identified contamination.

The following confusion matrix shows the results obtained on the test set mentioned before
$$\begin{bmatrix}
  9288&\hspace{-0.3cm}{(TN)}&1607&\hspace{-0.3cm}(FP)  \\
   3&\hspace{-0.3cm}{(FN)}& 110&\hspace{-0.3cm}{(TP)}  \\
\end{bmatrix}$$
We can see that the CNN classifier successfully does its job, reducing the rate of false positives from 58\% to less than 15\%. This reduction comes with a small cost,  the increase of the false negatives of a value of one.

\section{Conclusions}
\label{conclusions}
In the textile production the quality of the products is controlled in all the various steps of the process. 
This is in response to the increasing concern of the consumers in favor of sustainability of the processes that transform raw goods in apparel items. Besides attributes  like  the  standard  of  fibers,  yarns,  fabric  construction,  colour, shade, surface designs, fasteners and other accessories reliability, the finite product has to be free of intrusions, that are a clue of all the various production activities. These contaminations have to be detected, so that the item can be fixed before packaging is applied. Such a task is usually accomplished on X-ray images by expert personnel, but the activity is prone to mistakes, being highly repetitive and demanding from the attention point of view. 

In this paper an automatic procedure is developed to detect contaminations in apparel items, passed in an X-ray scanner. The system is composed of two subsystems: a filter and a classifier. The contaminations are very small objects immersed in a dense cloud of  background artefacts produced by the X-ray technology, therefore a general machine learning technique would easily fail in classifying items with contaminations from those free of them, by processing the whole X-ray image. Therefore the domain of the classifier must be reduced to a small neighbour of a possible intrusion. Identifying potential contamination is the task accomplished by the filter. By a multi-threshold approach, the filter is able to rapidly analyse an X-ray image and isolate a finite number of small objects that are possible intrusions to be further checked by the classifier. The filter subsystem was tuned on a training set of manually annotated images, and tested on a set of 11008 images; it reached good performances in terms of false negative rate ($<2\%$ of non detected contaminations) but with a false positive rate too high ($58\%$). The CNN classifier then operated on just the possible intrusions selected by the filter, and decreased the false positive rate to less than $15\%$, while mantaining the false negative rate below 3\%.
The procedure is quite fast, the response being within four seconds. The detection procedure can be easily updated as the production develops and may show intrusions of different kind or shape from those modelled.

\bibliographystyle{plain}
\bibliography{bibliography}
\end{document}